\documentclass[jou]{apa7}

\usepackage[american]{babel}
\usepackage{amssymb, amsmath, amscd, graphicx, fullpage, epstopdf, appendix, hyperref, color, subfig, setspace, tensor, url, verbatim, tabto, algorithm, algorithmic}

\usepackage{csquotes}
\usepackage[style=apa,sortcites=true,sorting=nyt,backend=biber]{biblatex}
\DeclareLanguageMapping{american}{american-apa}
\title{A Technique to Create Weaker Abstract Board Game Agents via Reinforcement Learning}

\author{Peter Jamieson and  Indrima Upadhyay}
\affiliation{Department of Electrical and Computer Engineering; Miami University}

\leftheader{Jamieson}

\abstract{Board games, with the exception of solo games, need at least one other player to play.  Because of this, we created Artificial Intelligent (AI) agents to play against us when an opponent is missing.  These AI agents are created in a number of ways, but one challenge with these agents is that an agent can have superior ability compared to us.

In this work, we describe how to create weaker AI agents that play board games.  We use Tic-Tac-Toe, Nine-Men’s Morris, and Mancala, and our technique uses a Reinforcement Learning model where an agent uses the Q-learning algorithm to learn these games.  We show how these agents can learn to play the board game perfectly, and we then describe our approach to making weaker versions of these agents.  Finally, we provide a methodology to compare AI agents.}


\begin{document}
\maketitle

\section{Introduction}

Reinforcement Learning  (RL) is a powerful model to create learning agents for complex problems. It is a research sub-area in the broader field of artificial intelligence (AI) where, in RL, an agent attempts to maximize the total reward for its actions in an uncertain environment. The application of RL in board games is interesting because board games present simplified spaces where we can test and observe a decision-based agent.  Also, board games provide a competitive space to compare different AI techniques all within a limited state-space complexity \textcite{Andrade2004-ANDOAO-3}. With a deeper understanding of the performance of RL game agents with respect to changing state-space complexity, we are able to evaluate AI techniques and provide insight on how to employ these agents in the broader board game market.  This includes providing a range of quality AI agents that are challenging for players at all levels of skill, which allows a human player to improve, beat, and have fun playing against.

Our goal is to explore the quantitative difference in the performance of RL-based game-playing agents with changing state-space complexity of abstract board games and to provide a methodology on how to create weaker agents from this learning process. For this purpose, our game-playing agents are designed using a Q-learning algorithm for three different games with different state-space complexities - Tic-Tac-Toe, Nine-Men's Morris, and Mancala. Each agent is trained against a range of AI agents (Min-Max, Q-learning, and Random) to get a better understanding of how quickly we find convergence to a solved state when training a Q-agent for a particular game.

In this paper, we will show our method to create weakened agents that will be compared against other AI agents.  Typically, for board games with much higher state-space complexity, such as chess, an agent is weakened by giving the agent less computational time to explore and evaluate the tree of possible moves.  However, in the case of games learned within Q-learning, the full matrix of good moves is known instantaneously, so weakening the agent computation time is not a relevant approach.  Instead, we use the learning process of the Q-learning algorithm playing against its mirror twin and snapshot the agents using these to create weaker agents. 

The contributions of this work are:
\begin{itemize}
    \item An analysis of training a Q-learning RL agent in terms of finding convergence for solved zero-sum abstract board games and evaluating training speed based on using a mirror of itself versus a Min-Max agent and a random agent.  This includes how to select parameters in the Q-learning algorithm.
    
    \item A method to create lower quality Q-learning-based agents to allow human players to find a competitive agent that can provide them challenging opponents that are not too hard to beat and possible to learn from. 
    
    \item A method to create a round-robin tournament to evaluate the quality of agents, which includes a method to determine how many games must be played between opponents to find stable results.
\end{itemize}

The results of these contributions provide other RL researchers with methods to properly evaluate RL agents and use Q-learning-based agents in the future for virtual game-playing agents.


\section{Background}
\label{B}

Board games are fixed environments that are good for experimenting with and testing various RL algorithms and agent-based learning techniques. RL has achieved human-level performance playing Atari games \textcite{Mnih13atari}. Temporal difference learning was used to create TD-Gammon, a game-learning program, which proved to be the world’s best player of Backgammon\textcite{Tesauro95acm}. RL is used by IBM’s Watson, in 2011, to make strategic decisions in Jeopardy!\textcite{Ferrucci12tiw}.

In this section, we provide details of the Ludii General Game System which is a platform on which we implement our agents.  Next, we look at different ways of measuring the complexity of an abstract board game. We describe details of our three games used in this work, the basics of RL, and the basics of our Q-learning and Min-Max algorithm.

All of our system and software has been released on Github and can be downloaded at: \url{www.github.com/indrima24} noting that this is a research repository and the software is not supported or detailed in usage.

\subsection{Ludii General Game System}
\label{lGGs}

The main objective of a General Game Playing (GGP) system is to provide a virtual environment that can implement various games \textcite{Genesereth05ggp2}. A GGP works by assuming that players and agents are not tightly coupled to a game, and therefore, this means game descriptions are similar to how game instructions are written. The logic description of games and their rules is known as a Game Definition Language (GDL) \textcite{Love06gpgl}. In recent years, there have been many advances in GGPs \textcite{Swiechowski15GGP}. 

Ludii General Game System, the GGP used in this research, is created by Digital Ludeme Project (DLP). DLP is Maastricht University’s GGP for reconstruction and analysis of numerous traditional and modern strategic abstract games, where an abstract game is defined as any game where the theme of the game is not important in terms of the game-playing experience.

The DLP facilitates the work of game designers, historians, and educators \textcite{Browne18mtag} and allows us to do our research on AI agents similar to other General Game Playing (GGP) systems such as FLUXPLAYER \textcite{Schiffel2007a}, which is based on FLUX, a high-level programming system for agents. 

Another GGP language, called Regular Board games (RBG), focuses on describing finite deterministic turn-based games with perfect information \textcite {DBLP:journals/corr/KowalskiSS17}. Ludii researchers believe DLP is superior in comparison with the performance of the Regular Board games (RBG) system, and have demonstrated that DLP is more general, extendable, and has other qualitative facets not found in RBG \textcite{Piette19ludii}.

\subsection{Game Complexity}

Combinatorial game theory has several ways of measuring game complexity \textcite{Demaine01a}. Game complexity is a quantitative measure of the game, and differs from game difficulty which is related to qualitative strategic elements. This means that a more complex game does not necessarily mean that it will be a more difficult one to play. 

\subsubsection{State-Space Complexity}

\textcite{5fc9fc423dcb4420b49effb13672c6c6} describes the state-space complexity of a game as the total count of legal game positions accessible from the starting point of the game. It is the number of all possible legal positions with the constraints being the initial position of any particular game \textcite{Piette19ludiiGGSintro}. Each time any of the pieces makes a legal move, a new board state is conceived; the sum total of all of which ends up as the game’s state-space complexity. 

For example, in the game of Tic-Tac-Toe, there are three potential states for each of the nine spaces in a 3x3 matrix: a cross, a naught, or neither. Simply by calculating three to the power of nine we find that a Tic-Tac-Toe board has 19,863 unique states. This number, however, comprises a variety of illegal situations, such as a situation where there are five naughts and no crosses, or a situation where crosses and naught both form rows of three. If all of the illegal positions on the board are removed from this count a total of 5,478 states remain, and when all reflections and rotations of positions are deemed as the same, there are only 765 fundamentally distinct board states. Researchers tend to focus on the upper and lower bounds of state-space complexity when calculating them \textcite{IJCAI07bikram}. 

The upper bound is the number of all the potential configurations of pieces including illegal positions and initial setup. In this case, having a situation with five naught and no cross would be included.  The lower bound is a measure of the minimum state-space complexity with only legal positions.

Even for uncomplicated games, state-space complexity amounts to large numbers, and we describe state-space complexity with log to base 10. This means that Tic-Tac-Toe has a upper bound state-space complexity of {\bf{4}} ({$\left \lceil log_{10}\left ( 5478 \right )\right \rceil$}).

\subsubsection{Zero-Sum Games and Fully Solved Games}

For the following games of study, we note two characteristics of them as defined as both ``Zero-sum'' and ''Fully Solved Games''.

Zero-sum games are games where one agent's gain is equivalent to the other one's loss and the sum total of all gains subtracted from the total losses, should sum to zero. This means that whenever one player wins, the other player loses. 

Fully solved games are games whose outcome from any particular state can be correctly calculated for perfect play. The optimal policy for such games can be calculated \cite{van2002games}, and this include games such as Connect Four \cite{allen1989note} and Checkers \cite{schaeffer2007checkers}.

\subsection{Abstract Games of Study}

To explore quantitative differences in the performance of RL-based game playing agents with changing state-space complexity of abstract board games we use three different board games described next in order of increasing state-space complexity.

\subsubsection{Tic-Tac-Toe}

Tic-Tac-Toe, also known as  Xs and Os or noughts and crosses, is a two-player game, where two players take turns in filling X and O in a 3×3 grid. To win the game a player must succeed in placing three of their marks in a horizontal, vertical, or diagonal row. It is a solved game with a forced draw assuming best play from both players. The board size, for Tic-Tac-Toe, is 9 and the state-space complexity (as log to base 10) is 4.
\textcite{VANDENHERIK2002277}

\subsubsection{Nine-Men's Morris}

Nine-Men’s Morris, also known as cowboy checkers, has a board that is made up of a grid with twenty-four intersecting points. There are two players and both of them have nine pieces each. Each player actively attempts to form ``mills'' where a mill is three of a player’s own pieces lined horizontally or vertically. This grants the player an action to eliminate an opponent's piece from the game. To win a player must either reduce the opponent to two pieces so that they cannot create any mills or leave them with no possible legal move. The board size, for nine men’s Morris, is 24 and the state-space complexity (as log to base 10) is 10 \textcite{7080922}.

\subsubsection{Mancala}

Mancala \textcite{Jones13smg} includes a board and 24 seeds. A rectangular board has 6 pits lined up consecutively on the north (player one's) and south (player two's), called houses, each of which contains 4 seeds each at beginning of the game. There is also a big pit at each end for each player (West is player one's and East is player two's), called the bank or store. Seeds are sown alternatively by each player. When it is a particular player’s turn they must remove all of the seeds from any one of their houses and move the seeds, dropping one seed at a time in each house except the opponent’s bank in a counter-clockwise direction. If the last sown seed lands in one of the player-owned empty houses and the opponent’s adjacent house contain seeds, both the player’s last seed and the opposition's seeds are put into the player's bank. If the last sown seed is in the player’s bank, they get an extra turn. The game ends when a player has no seeds in any of their houses, and the object of the game is to capture more seeds in your bank and houses when the game ends. The state-space complexity of Mancala is 13 \textcite{6633628}.

\subsection{Reinforcement Learning for Game Playing Agents}

An RL framework trains an agent’s behavior by making actions and using a reward function ({$R$}) when the agent performs well in an environment that is comprised of different states. For RL-based game-playing agents, the environment is a particular game, which has well-defined actions, and the agent has to decide which action must be selected to, eventually, win the game.  Moreover, the encouragement to learn new and good strategies stems from the result of agents getting rewards for winning a game.

Finite Markov Decision Process (MDP) \textcite{Ferns04MDP} is used for RL where an MDP requires:
\begin{enumerate}
\item{A finite set of actions (potential new positions in the game)}
\item{A finite set of states (a distinct arrangement on the board of a game)}
\item{A reward function that returns a value based on performing a specific action in a given state. This means that if the agent performs a particular action that results in finally winning the game from a given state, then {$R$} = 1. Otherwise, if the agent blunders and picks the erroneous action leading to eventually losing the game then {$R$} = -1. Else, in a situation where none of the above happens, the reward is {$R$} = 0.}
\item{ A transition function to provide the probability of moving from an existing state to a specific new state when performing a certain action.}
\end{enumerate}

The necessity for a robust transition function rises in situations where there is an uncertainty associated with an action in regards to eventually leading up to a certain desired result.  Finite states, actions, and rewards make up a finite MDP.  The framework of finite MDP acts as an aid in formalizing any problem so that actions, depending on a current state, can be easily recognized and thereby used to maximize the agent’s total reward during a game.

\begin{table}[ht!]
\caption{Examples of Tic-Tac-Toe states and Probability of Winning}
\label{tab:tictactoe}
\centering
\begin{tabular}{c|c}
\hline
Game State & Win Probability \\\hline
\hline
\includegraphics[width=1.3in]{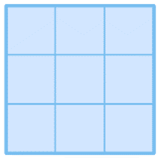}
 & 0.5 \\
\includegraphics[width=1.3in]{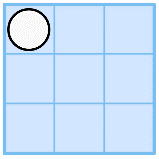}
 & 0.5 \\
 \includegraphics[width=1.3in]{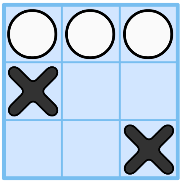}
 & 1(WIN) \\
  \includegraphics[width=1.3in]{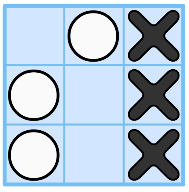}
 & 0(LOSS) \\
   \includegraphics[width=1.3in]{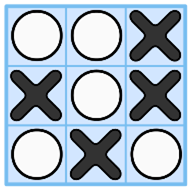}
 & 0(LOSS) \\
\hline
\end{tabular}
\end{table}

To understand the RL approach, consider a game of Tic-Tac-Toe and assume an imperfect opponent, i.e. an opponent that can lose or make mistakes. Table \ref{tab:tictactoe} shows various states and each entry in the table has an estimate of the probability of a win from that particular state.  This signifies the value of that state. Assuming the agent play O’s, estimating probability of winning is initialized as V[s = a state with three O’s in a row] = 1, V[s = a state with three X’s in a row] = 0, V[s = all draw states] = 0. 

Many games are played to train, and the subsequent moves at any point are chosen by looking ahead one step. The next state with the highest estimated probability of winning, a greedy move, is generally picked.  This determines the learning rule for the RL-based game-playing agent.

There are two main types of RL algorithms: 
\begin{enumerate}
    \item {Model-based RL algorithms: These try to use an experience to form an internal model of the transitions and immediate outcomes in the environment and then choose the optimal policy based on its learned model.  To do this a model can be learned by running a base policy, like a random policy or any educated policy, while observing the trajectory is observed to plan through the control function ($f(s,a)$) to choose the optimal actions.}
    \item {Model-free RL algorithm: These use an experience to directly update either state/ action values or policies, or both, which can then attain the identical optimal behavior but without the use of a world model or estimation, meaning the agent uses trial-and-error methods for achieving an optimal policy.  Q-learning is a model-free RL algorithm. It learns the action-value function ({$Q(s, a)$}), which is nothing but a scalar value assigned of an action $a$ given the state $s$.}
\end{enumerate}

For our research, we use a Q-learning algorithm because model-based methods are generally limited only to specific types of tasks.

Q-learning follows value-based learning. It is an RL algorithm used to learn the value of an action in a particular state \textcite{Hagen00q}.  The Q-learning algorithm operates without an understanding of the environment, and the basis of Q-learning is simply updating the Q-value, which is an estimation of how good performing an action would be in a given state, thereby, figuring out how to navigate an unknown environment.  We will describe this algorithm in more detail in the following section. 

\subsubsection{Min-Max Algorithm in the Context of Games}

The Min-Max algorithm minimizes the loss in a worst-case scenario situation. It finds the best possible move by modeling all possible continuations from a given position when the state of the board is provided and chooses the move that is best for the player. The best possible move is the one with the best outcome with the assumption that the player always makes the move that maximizes their reward value for it and the opponent always makes the move that maximizes the reward value for it, therefore, minimizing the game value for the player \textcite{doi:10.1137/1.9781611973082.20}.

For a two-player zero-sum game, the solution given by Min-Max algorithm is the same as the Nash equilibrium \textcite{yang11}. This implies that for any two-person, zero-sum game with a finite number of strategies, there exists a value V, and a mixed strategy for each player, such that Player 1's strategy makes sure that it achieves a payoff of V regardless of Player 2's strategy, and Player 2 ensures themselves a payoff of -V. 

We use two types of algorithm classes of the Min-Max algorithm in this work:
\begin{enumerate}
    \item Deterministic Min-Max agent: If there are more than one move with the exact same best scores in a given position, a deterministic Min-Max agent will always choose the same move.
    \item Non-deterministic Min-Max agent: If there are more than one move with the exact same best scores in a given position, a non-deterministic Min-Max agent will choose a move in a stochastic manner.
\end{enumerate}

\section{Analysis of Q-learning RL agents}
\label{C}

In this section, our goal is to analyze the training of a Q-learning-based game-playing agent with respect to the abstract games - Tic-Tac-Toe, Nine Men's Morris, and Mancala. It is important to note that Q-learning is a model-free algorithm, implying that it requires no model of the environment. This makes Q-Learning capable of dealing with stochastic transitions and rewards, with no need for adaptations and alterations.  We analyze how to train a Q-learning agent by using different training partners.  Specifically, we test how the training works using a twin of itself and both deterministic and non-deterministic Min-Max agents as the teaching agent. 

We report the number of episodes needed when training is done. An episode is the complete playing of a game. We, also, explore how quickly our agents train to converge under these different training opponents.  

We also provide a method to create weakened agents.  These agents are useful as they help learning humans play against opponents that make errors.  
 
\subsection{Details of the Q-learning Algorithm}

Q-learning is a model-free RL algorithm, which can be seen as a method of asynchronous dynamic programming (DP).  The Q-learning policy can be described as a function {$Q(s,a)$} that is maximized at a given state {$s$} when the agent performs an action {$a$}. Here {$Q$} is the expected reward the agent will get at the end of the game on performing a particular action {$a$} at specific state {$s$}. This acts as an encouragement for the agent to learn and pick the actions that maximize {$Q$} as it wants to increase the reward \textcite{Lorentz_ml}. The computation that forms the foundation of Q-Learning is: 
\begin{equation}
a_{best} =arg  \; max_{ a \in A} \; Q(s,a)
\end{equation}

To successfully calculate $Q(s, a)$, all probable pairs of states and actions have to be looked into by the agent while obtaining feedback from the reward function $R(s, a)$. $Q(s, a)$ can be revised repetitively by making the agent play several games against itself or an opponent. The equation used to update the value of $Q$ is based on the well-known equation within RL called the Bellman equation \textcite{KIUMARSI20141167}
The equation to update Q values is:

\begin{equation}
\label{Qeq}
Q(s, a)_{n} \leftarrow (1- \alpha ) \cdot Q(s, a) + \alpha \cdot (R(s, a)+ \gamma \;max_{a \in A}\;Q ( \hat{s}, \hat{a})
 \end{equation}
 
Here, $a$ is the action being performed in the current state $s$, the new state reached after performing action $a$ is $\hat{s}$ and $\hat{a}$ is the best possible action in the state $\hat{s}$, $\alpha$ is the learning factor, and $\gamma$ is the discount factor. 

Using a few experimental trials and consulting the literature (\textcite{MAHADEVAN1994164} \textcite{Wang18mcqlfggp}, \textcite{evendar03a} \textcite{Mahadevan94mlp}), we set $\gamma$ = 0.9 and $\alpha$ = 0.4 in all of our experiments.

\subsection{Training Methodology}

\begin{algorithm}
\caption{Pseudo-code for the training phase to learn the values of $Q(s, a)$}
\label{al}
\begin{algorithmic}[1]
\STATE Initialize: Q(s,a) = 0,
\STATE Starting state s,
\STATE Starting player P,
\STATE Number of games numGames;

\FOR {(t = 1; t <= numGames; t++)}
    \STATE Initialize: Q(s,a) = 0,
    \STATE With probability $\epsilon$: P picks random action $a$,
    \STATE Otherwise, picks an action a that maximizes $Q(s,a)$
    \STATE if(result == win) then reward $R(s,a)$ = winValue
    \STATE if(result == loss) then reward $R(s,a)$ = lossValue
    \STATE if(result == draw) then reward $R(s,a)$ = drawValue
    \STATE while(!starting state)
    \STATE Observe state ŝ and reward $R(s,a)$
    \STATE Update $Q(s,a)_{new} \leftarrow (1- \alpha ).Q(s,a) + \alpha .(R(s,a)+ \gamma \;max_{a \in A}\;Q$( ŝ, â)
 \ENDFOR
\STATE Switch turn, P = second player
\end{algorithmic}
\end{algorithm}

To train a Q-learning agent in an RL-based framework it needs to learn the values of $Q(s, a)$. This training is done by using two agents to play against each other including the possibility of having the learning agent play against itself.  We will call the agent that is being trained the ``Learning Agent'' and the agent that is being used to train the agent the ``Teaching Agent''.  

The probability of an agent choosing a random action is given by $\epsilon$, otherwise, it will choose to perform the best-known action according to $Q(s, a)$. This probabilistic control allows a learning agent to sometimes explore new actions and at other times exploit the information that it has already learned. The pseudo-code for our training phase is shown in algorithm \ref{al}. 

To understand how to get good Q values, assume that a table is created where the Q value for every potential state and move is stored.  Tabulated Q values might be calculated in a manner as explained below for an example in Tic-Tac-Toe.

\begin{figure*}[ht]
\centering
\includegraphics[width=4in]{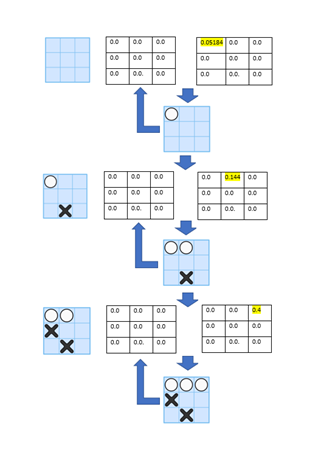}
\caption{Illustration of how the Q-agent trains.}
\label{pic7}
\end{figure*}

Figure~\ref{pic7} shows an example set of steps for a Tic-Tac-Toe game where the learning agent plays with a cross and is being trained against a teaching agent (doesn't matter if human or RL agent) playing naught that in this example is a random agent. Since O’s last move leads to a win, that action is rewarded with +1. The Q-value of that decision becomes 0 + $\alpha$* 1, since $\alpha$ =0.4, the Q value becomes 0.4. This value is used to update the previous action in the game history. Since the moves before this move do not end the game, there is no direct reward. Also since the discount factor, $\gamma$, is set to 0.9. And the maximum Q-value for the next state is 0.4, so we update Q-value to 0.144. Going one more state back in history, the starting position is reached. The maximum Q-value for the next state is 0.144, so the new Q-value for the first move turns out to be 0.05184. 

To train a Q-agent, the process is repeated for many episodes. The number of iterations, denoted by $N$ in algorithm \ref{al}, must be fairly substantial and we will show results n the next section of how we determine $N$ experimentally. The value of $N$ will change depending on the different teaching agents, and the state-space complexity of the board game being trained for as we will show later in this section.

Once the value of the Q function is known, a game is played simply by observing the Q values for all the moves possible in the current state and the move with the highest Q value is picked. In situations where there is more than one possible move with the same highest value, a move is chosen randomly amongst the options available. Having the highest value means that this move is the best move in a given situation. 

\subsection{Stopping Criteria For Training}
\label{SCT}

In this section, we answer the following questions with respect to training time: How many generations does a learning agent need to train for before the algorithm converges on the best solution?  Also, what is the ``best solution''?  How do we decide if the agent is good enough?  These questions are in relation to how we find, $N$ (the number of training iterations) in algorithm \ref{al}.

\subsubsection{Definitions: Convergence until Stable versus Convergence until Fully Solved}
\label{Conv}

Before we delve into training our agents it is important to understand the two conditions of convergence that we are working with. 

In our research, ``Convergence until Fully Solved'' (Convergence-FS) happens once a learning agent appears to have achieved perfect play, that is it meets our stopping criteria against a strong opponent.  In our methodology, we perform an extra 5000 trials of training to ensure that the agent hasn't just gotten "lucky", and that training has truly Converged-FS.  

Convergence until stability (Convergence-ST) is the number of epochs an agent requires to initially reach a win/draw to loss ratio which remains constant for at least the next 20 trials with a fluctuation of one.

This means that Convergence-ST is a defined stopping mechanism, i.e. understanding it gives us guidelines on observing the number of games required to achieve a win/draw to loss ratio that can be categorized as stable despite minimal perturbations.

For the learning agent, if the teaching agent is too weak, then it might attain Convergence-ST, however, still remain unable to achieve Convergence-FS .

We show our results for {$N$}, the number of training epochs required for the algorithm to converge until fully solved, in the following sections for each of the three games.

\subsection{Training Tic-Tac-Toe Agent Against Different Teaching Agents}
\label{SCT1}

To see how a Q-learning agent trains, a learning agent for the game of Tic-Tac-Toe trains against the following teaching agents:
\begin{enumerate}
 \item{Q-learning-based game-playing agent}: The twin of the learning agent, which itself is changing in each epoch.
 \item{Non-deterministic Min-Max agent}: A Min-Max agent designed by us such that it randomly chooses a move, if there is more than one move with the exact same best scores in a given position.
 \item{Deterministic Min-Max agent}: A Min-Max agent designed by us such that it always chooses the same move if there is more than one move with the exact same best scores in a given position.
 \item{Random agent}: An agent that makes its move purely randomly, i.e., it looks for a random empty spot on the board and puts its piece there.
\end {enumerate}

 \begin{figure}[ht!]
\centering
\includegraphics[scale=0.75]{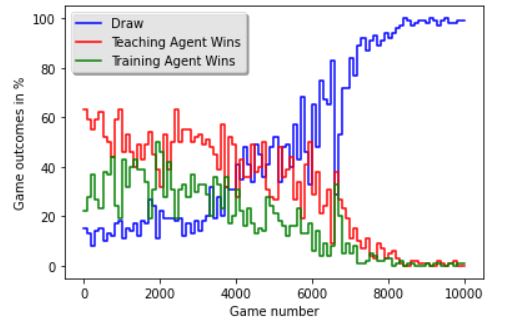}
\caption{ Q-learning-based game-playing agent for Tic-Tac-Toe learning by playing against a Q-learning-based player}
\label{pic8}
\end{figure}

The Q-learning learning agent trained with itself as the teaching agent is shown in Figure \ref{pic8}, and reaches convergence-FS point after approximately eight thousand games.  Note that in this figure, the curve in green denotes games that end in player 1 winning, which is the learning agent (the Q-learning agent being trained), the green line denotes games that end in player 2 (the teaching agent) winning, and blue line denotes games that end in a draw. In this figure, the x-axis shows the training game number proceeding from 0 to the convergence-FS number plus 5000 where a game number is equivalent to a training generation, and the y-axis is a cumulative game outcomes in percentage. Note that after each training session, we play 100 games for each ``Game Number'' that give us an outcome in \% on the y-axis.  For example, for ``Game Number'' 2000 on the x-axis in Figure \ref{pic8}, the number of draws over 100 games is 10 (in blue), Teaching agent wins 40 (in red), and the Q-learning agent wins 50 (in green). 

\begin{figure}[ht!]
\centering
\includegraphics[scale=0.75]{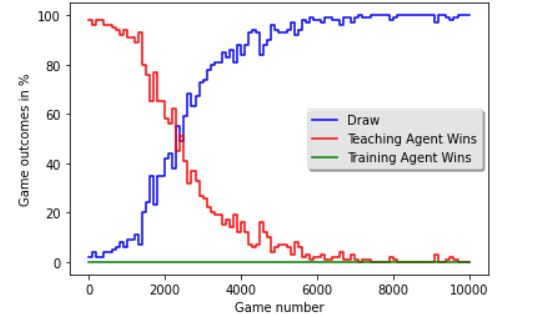}
\caption{Q-learning-based game-playing agent Tic-Tac-Toe learning by playing against non-deterministic Min-Max player.}
\label{pic9}
\end{figure}

Figure \ref{pic9} shows the learning agent (Q-learning) trained against a non-deterministic Min-Max teaching agent, and the learning agent converges-FS after approximately six thousand games, and this graph follows the same conventions as Figure \ref{pic8}. Note that the learning agent ( Q-learning agent) has a straight line parallel to the x-axis, and this is because for the scope of this research in the game of Tic-Tac-Toe a draw is as good as a win condition and the learning agent pursues a draw.  

We also perform the same training for the other two teaching algorithms.  We observe how a learning agent trains with a random teaching agent that even after fifty thousand games, our agent was still losing games.  This result makes sense since the learning agent can not find good strategies (unless by random chance) from a teaching agent that cannot play the game, at least, sufficiently well.  However, the learning agent does perform well in terms of winning, but the agent never converges from our stopping criteria definition of convergence-FS.

\begin{table}[ht!]
\centering
\caption{Summary of Training until Convergence for Tic-Tac-Toe with Different Teaching Agents}
\begin{tabular}{c|c} 
 \hline
 Teaching Agent & $N$ until Convergence \\  
 \hline
 Q-learning Twin & 8000 \\ 
 \hline
 Non-deterministic Min-Max & 6000 \\
 \hline
 Deterministic Min-Max & 200 \\
  \hline
 Random & - \\
  \hline
 Fully Converged Q-learning & 300 \\
  \hline
\end{tabular}
\label{table2}
\end{table}

Table \ref{table2} shows the convergence-FS results of {$N$} for all of the different scenarios.  Column 1 lists the teaching agent and column 2 shows the approximate value when the learning agent (Q-learning in all cases) converges based on our stopping criteria. We can see that the Deterministic Min-Max agent is the best teaching agent as it provides clearer guidance to the learning agent, but we also see that using a converged Q-learning agent has similar results.  In the cases when existing teaching agents aren't available, then using the twinned Q-learning is a viable means to train an agent, however, to accelerate the training process of a known game with finite states Q-learning-based player should be trained by making it play against a higher quality agent. Currently, we have no hypothesis on why the Deterministic Min-Max agent is better than the converged Q-learning agent.

\subsection{Training Nine-Men's Morris Agent Against Different Teaching Agents}

The Q-learning-based player for Nine-Men's Morris, a game with a state-space complexity of 10, was also trained with different teaching agents.  Note, that we no longer use a random agent as this type of training was shown to be not useful in Tic-Tac-Toe.

\begin{table}[ht!]
\centering
\caption{Summary of Training until Convergence for Nine-Men's Morris with Different Teaching Agents}
\begin{tabular}{c|c} 
 \hline
 Teaching Agent & $N$ until Convergence \\  
 \hline
 Q-learning Twin & 16000 \\ 
 \hline
 Non-deterministic Min-Max & 7000 \\
 \hline
 Deterministic Min-Max & 2000 \\
  \hline
 Fully Converged Q-learning & 3000 \\
  \hline
\end{tabular}
\label{table4}
\end{table}

Table \ref{table4} shows the convergence results for all of the different teaching agents.  Column 1 lists the teaching agent and column 2 shows the approximate value when the learning agent (Q-learning in all cases) converges based on our stopping criteria.  With similar results to Tic-Tac-Toe training, we can see that the Deterministic Min-Max agent is the best teaching agent as it provides clearer guidance to the learning agent.  Still, for cases when existing teaching agents aren't available, then using the twinned Q-learning is a viable means to train an agent.  Also, notice how the increase in state-space complexity results in longer training times.

\subsection{Training Mancala Q-learning Agent against Different Teaching Agents}

\begin{table}[ht!]
\centering
\caption{Summary of Training until Convergence for Mancala with Different Teaching Agents}
\begin{tabular}{c|c} 
 \hline
 Teaching Agent & $N$ until Convergence \\  
 \hline
 Q-learning & 32000 \\ 
 \hline
 Non-deterministic Min-Max & 17000 \\
 \hline
 Deterministic Min-Max & 9000 \\
  \hline
 Fully Converged Q-learning & 12000 \\
  \hline
\end{tabular}
\label{table3}
\end{table}

Table \ref{table3} again shows similar results to both the previous games in that a teaching agent that is a Deterministic Min-Max provides the most efficient training of our learning agent.  The results follow a similar pattern to the previous games, and we can see that as State-space complexity increases so does the size of {$N$} increase for each of the games.   

\subsection{Snapshots for Creating Lower Quality Agents}
\label{secSnap}

As described earlier, the goal of a game agent is not necessarily always to create the ``best'' agent but to provide humans with artificial opponents who are competitively matched to the humans' current capabilities.  In this section, we provide a methodology to create these weaker agents via the idea of taking snapshots while training.

Our methodology suggests that we maintain data of the learning agent throughout the training process.  This can be achieved by saving the Q matrix to a file as the training proceeds - these files are then snapshots of the trained agent at some point in time on the training scale.  Next, once we have met our stopping criteria condition we then have a point in time that we call by the parameter $N$ in algorithm \ref{al} as convergence-FS.  We can then use the snapshot at time $N$ and call it Q-100, which represents the Q-learning agent as 100\% trained.  Using this approach we can extract other snapshots to create other Q-percent agents where a 50\% agent uses the snapshot file $N/2$ as its learned state.

The next question is which teaching agent should we use as the teaching agent for creating our snapshots.  In this work, we have selected the Q-learning twin as the teaching agent of choice for our snapshot methodology.  The reason for this is that the convergence graphs for this teaching agent (Such as the stages depicted in Figure \ref{pic7}) have the most gradual steps in the training process, and we hypothesize that the gradual nature is best for our snapshot method.  However, we do not have conclusive data on if this approach is best, and we leave this as future work where an experiment would need to be created where different approaches to snapshots could be compared either with human opponents or some measured comparison.

Given our methodology, since the Q-learning agent converges against itself after 8000 Tic-Tac-Toe games, we define $N=8000$ and define a ten percent agent for Tic-Tac-Toe, denoted ad Q-T-10, as the Q-matrix of the agent at 10 percent of 8000 which is 800 epochs into training. 

In the following section, we will use our snapshot agents in the evaluation results to observe how these opponents compare against the converged-FS agents.

\section{Comparing Agents}
\label{E}

The next question with RL-based training of agents that we address is how can we compare a range of agents against each other?  In particular, we want to evaluate our Q-learning agent and its snapshots of weakened Q-learning agents to existing algorithms to understand the relationship between the quality of these agents.  More importantly, we present a methodology on how to evaluate a range of agents for abstract board games.  We provide details and results of this approach for evaluation noting that it can be modified and used by other researchers for other agent comparison needs.

To evaluate the performance of AI agents the first step is to define the agents' intended purpose and what constitutes success.  In this work, the intended purpose is to see how well the agents perform in an abstract game with changing state-space complexity of competitive multi-player games against a range of agents.  In the space we are examining, a key property of each of the games under investigation is that they are solved games (even though each game has a different state-space complexity).

In previous research, a number of methods have been proposed to evaluate players/agents against one another.  We are dealing with multi-player games and the opponents are not always constant. Thus, one popular approach in these spaces is an aggregated relative performance \textcite{Volz20Benchmark} as a popular performance metric that might suffice our need. There are two different ways in which aggregating the results can help in quantifying the performance of the agent.  These are achieved by:
\begin{enumerate}
    \item {computing the win rates by distributing points per match-up (this is used in many sports leagues) \textcite{Volz20Benchmark}}
    \item {using iterative measures such as player rating (for example, MMR in StarCraft II, ELO in chess, etc.)\textcite{Volz20Benchmark}}
\end{enumerate}

Since we don't have an established ranking system to rate the players/agents, we focus on computing the win rates for our evaluation method. The aim, here, is to run multiple matches of a game-playing agent against multiple agents and against itself. The number of matches won, lost or drawn for a particular agent for each of the three games - Tic-Tac-Toe, Nine-Men's Morris, and Mancala will be recorded and presented.

The first step, however, is to define what ``good'' is in terms of wins, draws, and losses.  Next, we describe a methodology to identify the number of games played between a set of opponents where that number of games will have a high likelihood of resulting in a deterministic result, meaning that the reported win, loss, and draw rate will only have small perturbations if the number of games is increased.  Finally, we use this found number of games and execute our round-robin tournament to compare the agents.  

We will treat each game by defining what is ``good'' for an agent, then we will find the necessary number of matches needed to be played in a round of the tournament, and then finally, we will provide results from the round-robin tournament.  For each of the three games (Tic-Tac-Toe, Nine Men's Morris, and Mancala) we will have a tournament with the following agents:
\begin{itemize}
    \item A fully converged Q-learning agent, Q-100
    \item Q-learning agent at 50 percent, Q-50
    \item Non-deterministic Min Max (Nd-Min-Max)
    \item Deterministic Min Max (D-Min-Max)
    \item Random (rand)
\end{itemize}

\subsection{Evaluating Tic-Tac-Toe}

\subsubsection{Defining a ``Good'' game agent for Tic-Tac-Toe}

For Tic-Tac-Toe a converged solved agent playing another agent results in a forced draw assuming best play from both players.  So to quantitatively decide an agent is ``good'' we would expect it to draw against another ``good'' agent.  However, against a weaker agent, we would expect that a ``good'' agent would win.  However, note during the training of the agent it is equally motivated to draw, and therefore, this tournament structure will reveal if one RL agent is superior to another if it has more wins against other agents.

\begin{table*}[ht!]
\centering
\caption{The baseline after 10000 games of Tic-Tac-Toe with Min-Max and Random agent as players}
\begin{tabular}{c|c| c c c} 
 \hline
 \bf{Player-1} & \bf{Player-2} & \bf{Player-1 wins} & \bf{Player-2 wins} & \bf{draws}\\  
 \hline
 Min-Max agent & Random agent & 99.5\% & 0\% & 0.5\%  \\ 
 \hline
 Random agent &  Min-Max agent & 0 \% & 80 \% & 20 \% \\
 \hline
 Min-Max agent & Min-Max Agent & 0 \% & 0 \% & 100 \% \\
 \hline
\end{tabular}
\label{table5}
\end{table*}

Table \ref{table5} shows some simple baseline results of ``good''.  In this experiment, we show that as player 1 a ``good'' agent should almost win all the time against a random agent.  The Min-Max agent as player 2 versus a random agent will win fewer games, approximately 80\%, and will draw the remainder of the games.  In general, the minimum requirement of an agent is drawing each game, and better agents will win more.  Second, we can see that depending on the agent's playing position (first or second to play) the results will differ.  Therefore, our full round-robin tournament will have agents for both player 1 and player 2.

\subsubsection{Method to find the Number of Games to find a Stable Result for Tic-Tac-Toe between Two Agents}

\begin{figure}[ht!]
\centering
\includegraphics[scale=0.75]{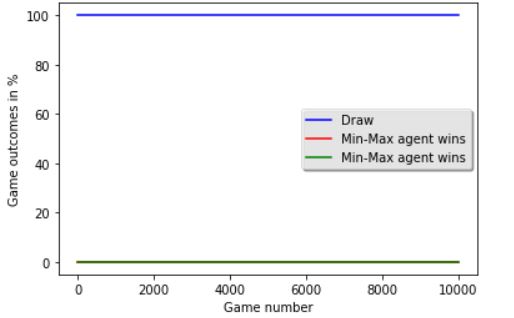}
\caption{10000 games between Min-Max - Min-Max players to establish baseline}
\label{chap4pic3}
\end{figure}


\begin{figure}[ht!]
\centering
\includegraphics[scale=0.75]{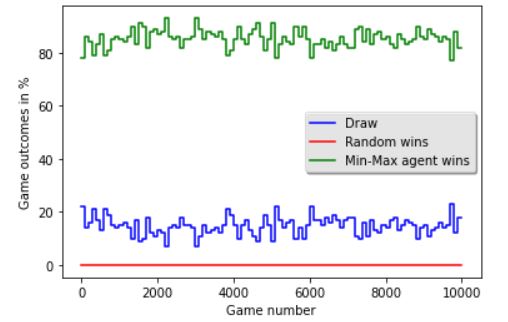}
\caption{10000 games between Random - Min-Max players to establish baseline}
\label{chap4pic2}
\end{figure}

Our setup for evaluation is a round-robin tournament, but the question remains, how many matches need to be played between opponents to find a stable result?  For example, two converged agents only need to play 1 game since they both will result in a draw as both agents are sufficiently ``good''; Figure~\ref{chap4pic3} shows the instant stability of this.  However, a random agent will never come to a stable result and will fluctuate in wins, draws, and losses.  This is shown in Figure~\ref{chap4pic2} where a Random agent vs. Min-Max agent.  In this graph, we can see how the Min-Max agent wins most of the time and draws many games, but a clearly stable space is not found.

The goal here is to find $N_{trialsTTT}$ which is the number of games played between opponents to find a stable result.  To do this, we will run a set of experiments and find $N_{trialsTTT}$ based on finding stable results under a number of conditions and take the maximum. 

We will start with a constant that is defined as $N_{1}$ and we make this 100 games since this number is computationally feasible for our experiments.  

We define the number of trials required for a ``stable'' win/draw to loss ratio as the number of games an agent requires to initially reach a win/draw to loss ratio which remains constant for at least the next 20 trials with a fluctuation of one (convergance-ST). 

\begin{table}[ht!]
\centering
\caption{Number of trials the Q-learning agent for Tic-Tac-Toe requires to reach a stable win/draw to loss ratio}
\begin{tabular}{c|c|c} 
 \hline
 Player-1 & Player-2 & Episodes to stability \\  
 \hline
 Q-20 & rand & $N_{T1a}$ = 35 \\ 
 \hline
 Q-20 & Nd-Min-Max & $N_{T1b}$ = 20 \\
 \hline
 Q-20 & D-Min-Max & $N_{T1c}$= 7 \\
 \hline
 Q-20 & Q-50 & $N_{T1d}$ = 27 \\
  \hline
 Q-20 & Q-100 & $N_{T1e}$ = 23  \\
  \hline
 Q-40 & rand & $N_{T2a}$ = 32 \\ 
 \hline
 Q-40 &  Nd-Min-Max & $N_{T2b}$ = 18 \\
 \hline
 Q-40 & D-Min-Max & $N_{T2c}$= 5 \\
 \hline
 Q-40 & Q-50 & $N_{T2d}$ = 25 \\
  \hline
 Q-40 & Q-100 & $N_{T2e}$ = 20 \\
  \hline
 Q-60 & rand & $N_{T3a}$ = 30 \\ 
 \hline
 Q-60 &  Nd-Min-Max & $N_{T3b}$ = 15 \\
 \hline
 Q-60 & D-Min-Max & $N_{T3c}$= 3 \\
 \hline
 Q-60 & Q-50 & $N_{T3d}$ = 23 \\
  \hline
 Q-60 & Q-100 & $N_{T3e}$ = 17 \\
  \hline
 Q-80 & rand & $N_{T4a}$ = 26 \\ 
 \hline
 Q-80 &  Nd-Min-Max & $N_{T4b}$ = 11 \\
 \hline
 Q-80 & D-Min-Max & $N_{T4c}$= 2 \\
 \hline
 Q-80 & Q-50 & $N_{T4d}$ = 20 \\
  \hline
 Q-80 & Q-100 & $N_{T4e}$ = 15 \\
  \hline
\end{tabular}
\label{tableX}
\end{table}

As fully converged trained agents will result in a ``stable'' rate of one, we use weaker agents, such as our snapshot agents, spaced at regular intervals (of 20\%) to find the total number of trials required. These findings are summarised in table \ref{tableX}.  

Now the number of trials for the tournament of Tic-Tac-Toe can be determined by the Maximum of this and the total number of trials required to set up a Tic-Tac-Toe tournament is thus, $N_{trialsTTT}$ = 100, which though greater than all of the results in the table is computationally feasible to run the tournament on.  

\subsubsection{Round-Robin Match tournament Results for the game of Tic-Tac-Toe}

Now that we have found out the total number of trials, $N_{trialsTTT}$ = 100, required for stable results, we perform and present the tournament results for all of the agents. Table \ref{tx} contains the summary of the tournament with column 1 being the agents acting as player 1 and row 1 listing the agents as player 2, where the players are:  Q-learning agent at 100 percent (Q-100),  Q-learning agent at 50 percent (Q-50), Deterministic Min-Max Agent ($M_{det}$), Non-deterministic Min-Max agent ($M_{Ndet}$), and random (R).  Each entry is in the form P1-wins:Draws:P1-losses where for example 2:94:4 means that player 1 won 2 games, drew 94 games, and lost 4 games. The last column shows the total points earned by a particular agent, where 3 points are given to each game won and 1 point is given to each game drawn.

\begin{table*}[ht!]
\centering
\caption{Tic-Tac-Toe tournament to look at win/draw rate for each of the players}
\begin{tabular}{c| c c c c c| c} 
 \bf{P1/P2} & \bf{Q-100} & \bf{Q-50} & \bf{$M_{det}$} & \bf{$M_{Ndet}$} & \bf{R} & \bf{Points} \\
 \hline
\bf{ Q-100} &  0:100:0 & 5:95:0 &  0:100:0 & 0:100:0 & 95:5:0 & 700\\
\bf{ Q-50} & 0:75:25 & 60:5:35 & 0:50:50 & 0:52:48 & 5:55:40 &  432\\
\bf{ $M_{det}$} & 0:100:0 & 10:90:0 & 0:100:0 & 0:100:0 & 100:0:0 & 720\\
\bf{ $M_{Ndet}$} & 0:100:0 & 5:95:0 & 0:100:0 & 0:100:0 & 90:10:0 & 690\\
\bf{ R} & 15:10:75 & 0:15:85 & 0:25:75 & 0:20:80 & 65:5:30 & 315\\
\end{tabular}
\label{tx}
\end{table*}

Table \ref{tx} shows that a fully converged Q-Learning (Q-100) agent is a "good" agent and performs almost on par with deterministic Min-Max agent and non-deterministic Min-Max agent. It can also be seen that while Q-learning agent at 50 percent (Q-50) performs much better than a random agent.
By looking at the total points earned by each of the players we can see that the deterministic min-max agent is the best agent while a fully converged Q-Learning agent is a close second. 

\subsection{Evaluating Nine Men's Morris}

\subsubsection{Defining ``Good'' game-playing agents for Nine Men's Morris}
\label{good2}

For Nine Men's Morris, a ``good'' solution is a draw \textcite{GassersSNMM96} assuming best play from both players.  So to quantitatively decide an agent is ``good'' we would expect it to draw against another ``good'' agent.  However, against a weaker agent, we would expect that a ``good'' agent would win.  However, the interesting idea of training is that the agent may be motivated to draw, and therefore, this tournament structure will reveal if one RL agent is superior to another if it wins against weaker agents.

\subsubsection{Round-Robin Match tournament Results for the game of Nine Men's Morris}

The total number of trials required to set up a Nine Men's Morris tournament is calculated the same way as Tic-Tac-Toe and has $N_{trialsTTT}$ = 100, which though greater than all of the results in the table is computationally feasible to run the experiments.  Also note, the max number has increased to 55 for this game.

Now that we have found out the total number of trials, $N_{trialsTTT}$ = 100, required for stable results, we perform and present the tournament results for all of the agents. Table \ref{tx2} contains the summary of the tournament with column 1 being player 1 and row 1 being player 2, where the players are:  Q-learning agent at 100 percent (Q-100),  Q-learning agent at 50 percent (Q-50), Deterministic Min-Max Agent ($M_{det}$), Non-deterministic Min-Max agent ($M_{Ndet}$), and random (R).  Each entry is in the form P1-wins:Draws:P1-losses where for example 2:94:4 means that player 1 won 2 games, drew 94 games, and lost 4 games. The last column shows the total points earned by a particular agent, where 3 points are given to each game won and 1 point is given to each game drawn.

\begin{table*}[ht!]
\centering
\caption{Nine Men's Morris tournament to look at win rate for each of the players}
\begin{tabular}{c | c c c c c | c} 
 \bf{P1/P2} & \bf{Q-100} & \bf{Q-50} & \bf{$M_{det}$} & \bf{$M_{Ndet}$} & \bf{R} & \bf{Points} \\
 \hline
\bf{ Q-100} & 0:100:0 & 3:97:0 & 0:100:0 &  0:100:0 &  92:8:0 & 690 \\
\bf{ Q-50} & 3:90:7 & 5:60:35  & 0:45:55 & 0:50:50 & 20:35:45 & 364\\
\bf{ $M_{det}$} & 0:100:0 & 8:92:0 &  0:100:0 & 0:100:0 & 92:8:0 & 700\\
\bf{ $M_{Ndet}$} & 0:100:0 & 5:95:0 & 0:100:0 & 0:100:0 & 94:6:0 & 698\\
\bf{ R} & 10:15:75 & 0:17:83 &  0:25:75 & 0:22:78 & 10:55:35 & 194\\
\end{tabular}
\label{tx2}
\end{table*}

Looking at the win rate of each of the players from table \ref{tx2} we can infer that fully converged Q-Learning agent is a "good" agent and performs on par with deterministic Min-Max agent and non-deterministic Min-Max agent. It can also be seen that while Q-learning agent at 50 percent (Q-50) performs much better than a random agent.

By looking at the total points earned by each of the players, where 3 points are given to each game won and 1 point is given to each game drawn, we can see that a deterministic min-max agent is the best agent while non-deterministic Min-Max and the fully converged Q-Learning agents are both following closely. 

\subsection{Evaluating Mancala}

\subsubsection{Round-Robin Match tournament Results for the game of Mancala}

Using the same approach, the total number of trials required to set up a Mancala tournament is , $N_{trialsM}$ = 115.  Also, note that this result shows how our initial {$N_1$} choice of 100 has been surpassed as the stability point have increased for this games increases state-space complexity.

Now that we have found out the total number of trials, $N_{trialsM}$ = 115, required for stable results, we perform and present the tournament results for all of the agents. Table \ref{tx} contains the summary of the tournament with column 1 being player 1 and row 1 being player 2, where the players are:  Q-learning agent at 100 percent (Q-100),  Q-learning agent at 50 percent (Q-50), Deterministic Min-Max Agent ($M_{det}$), Non-deterministic Min-Max agent ($M_{Ndet}$), and random (R).  Each entry is in the form P1-wins:Draws:P1-losses where for example 2:94:4 means that player 1 won 2 games, drew 94 games, and lost 4 games. The last column shows the total points earned by a particular agent, where 3 points are given to each game won and 1 point is given to each game drawn. 
\begin{table*}[ht!]
\centering
\caption{Mancala tournament to look at win/draw rate for each of the players}
\begin{tabular}{c| c c c c c| c} 
 \bf{P1/P2} & \bf{Q-100} & \bf{Q-50} & \bf{$M_{det}$} & \bf{$M_{Ndet}$} & \bf{R} & \bf{Points} \\
 \hline
 \bf{Q-100} &  115:0:0 & 115:0:0 &  115:0:0 & 115:0:0 & 110:5:0 & 1715\\
 \bf{Q-50} & 58:5:52 & 60:5:50 & 46:31:38 & 49:21:30 & 55:35:10 & 901\\
\bf{$M_{det}$} & 115:0:0 & 114:1:0 & 115:0:0 & 115:0:0 & 114:1:0 & 1721\\
\bf{ $M_{Ndet}$} & 115:0:0 & 114:1:0 & 115:0:0 & 115:0:0 & 112:3:0 & 1717\\
\bf{ R} & 35:0:80 & 54:15:46 & 37:29:49 & 39:50:26 & 45:15:55 & 739\\
\end{tabular}
\label{tx3}
\end{table*}
 
Table \ref{tx3} proves that a fully converged Q-Learning agent is a "good" agent and performs almost on par with deterministic Min-Max agent and non-deterministic Min-Max agent. It can also be seen that while Q-learning agent at 50 percent (Q-50) performs much better than a random agent.

By looking at the total points earned by each of the players we can see that the deterministic min-max agent is the best agent while the non-deterministic min-max agent is second.  Also, in this game, we see the strong impact of how important it is to go 1st for Mancala as even the random agent can do quite well going first. 

\section{Discussion and Conclusion}
\label{D}

In this paper, we described our results and methodologies for training abstract board game RL agents using a Q-learning algorithm.  We showed how the training itself can be done via a snapshot technique of the Q-learning matrix to create weaker agents that are useful for matching up to the level at which a learning human is at.  Finally, we provided a methodology to compare agents against each other in a round-robin tournament with a methodology on how many games need to be played in each round of the tournament.

For all the games used in this research the perfect play (the action or behavior of any game-playing agent) that leads to the best possible outcome for that player no matter the response by the opponent, is known as these games are solved. An interesting extension of our work could be to implement similar techniques for partially solved and unsolved games like Chess, Go, and Reversi.  Our hypothesis is that we could create weakened agents, but it would be difficult to quantify how good these agents are.
 
Additionally, since, we hypothesize that Q-learning will only be useful as an RL-based technique for a limited state-space complexity, therefore, it might be fascinating to see if our observations of the agents hold good for more computationally expensive games like Stratego (state-space complexity: 115), Twixt (state-space complexity: 140), Connect6 (state-space complexity: 170), etc. This approach might even be extended to observe the game-playing agents' behavior when playing infinite games, i.e., games that are played to keep the gameplay going. Infinite games exist by playing with boundaries and rules of the game, for example, Infinite Chess, Magic: The Gathering, etc. These games can be treated as continuous tasks, i.e., RL tasks which are not made of episodes but instead last forever.  The problem with Q-learning, however, is the memory space needed to store the Q matrix.

It should be noted that we created lower-quality agents by taking snapshots while training. It might also be interesting to explore other methods to create these weaker agents. Such methods may include, but are not limited to, training against weaker opponents, tuning the parameters $\alpha$, the learning factor, and $\gamma$, the discount factor, and $\epsilon$, the exploration parameter, in such a manner that Q-Learning converges prematurely (before actually reaching the optimal policy), or perturbing the Q-matrix once trained.

\printbibliography

@unpublished{Al,
		author={Alastair Smith},
		title={{Private Communications with Alastair Smith}},
		year=2009
	    }

@article{Andrade2004-ANDOAO-3,
	author = {G. {Andrade} and H {Santana} and A. {Furtado} and A.{Leit\~ao} and G. {Ramalho}},
	title = {Online Adaptation of Computer Games Agents: A Reinforcement Learning Approach},
	year = {2004},
	volume = {15},
	number = {2},
	journal = {Scientia}
}

@INPROCEEDINGS{Mnih13atari,
author={V. {Mnih} and K. {Kavukcuoglu} and  D. {Silver} and A. {Graves} and I. {Antonoglou} and D. {Wierstra} and M. {Riedmiller}},
booktitle={arxiv:1312.5602, NIPS Deep Learning Workshop},
title={Playing Atari with Deep Reinforcement Learning},
year={2013},
volume={},
number={},
pages={},
}

@INPROCEEDINGS{Genesereth05ggp2,
author={M. {Genesereth} and N. {Love} and B. {Pell}},
booktitle={AI Magazine},
title={General game playing: overview of the AAAI competition},
year={2005},
volume={26},
pages={62–72}
}

@INPROCEEDINGS{Love06gpgl,
author={N. {Love} and T. {Hinrichs} and M. {Genesereth}},
booktitle={Stanford Logic Group LG-2006-01, Computer Science Department, Stanford University, Stanford, Calif, USA},
title={General game playing: game description language specification},
year={2006}
}

@INPROCEEDINGS{Swiechowski15GGP,
author={M. {Świechowski} and H. {Park} and J. {Mańdziuk} and K. {Kim}},
booktitle={The Scientific World Journal},
title={Recent Advances in General Game Playing},
year={2015},
volume={2015},
number={},
pages={22}
}

@InProceedings{Demaine01a,
author="Demaine, Erik D.",
editor="Sgall, Ji{\v{r}}{\'i}
and Pultr, Ale{\v{s}}
and Kolman, Petr",
title="Playing Games with Algorithms: Algorithmic Combinatorial Game Theory",
booktitle="Mathematical Foundations of Computer Science 2001",
year="2001",
publisher="Springer Berlin Heidelberg",
address="Berlin, Heidelberg",
pages="18--33",
isbn="978-3-540-44683-5"
}

@inproceedings{Schiffel2007a,
  author = {Stephan Schiffel and Michael Thielscher},
  title = {Fluxplayer: A Successful General Game Player},
  booktitle = {Proceedings of the 22nd AAAI Conference on Artificial Intelligence (AAAI-07)},
  publisher = {AAAI Press},
  year = {2007},
  pages = {1191-1196},
  url = {http://www.fluxagent.org/download.php?file=07-SchiffelThielscher-AAAI.pdf}
}

@article{DBLP:journals/corr/KowalskiSS17,
  author    = {Jakub Kowalski and
               Jakub Sutowicz and
               Marek Szykula},
  title     = {Regular Boardgames},
  journal   = {CoRR},
  volume    = {abs/1706.02462},
  year      = {2017},
  url       = {http://arxiv.org/abs/1706.02462},
  archivePrefix = {arXiv},
  eprint    = {1706.02462},
  bibsource = {dblp computer science bibliography, https://dblp.org}
}

@phdthesis{5fc9fc423dcb4420b49effb13672c6c6,
title = "Searching for solutions in games and artificial intelligence",
author = "L.V. Allis",
year = "1994",
month = jan,
day = "1",
language = "English",
isbn = "9090074880",
publisher = "Rijksuniversiteit Limburg",
school = "Maastricht University",
}

@INPROCEEDINGS{Jones13smg,
author={B. {Jones} and L. {Taalman}  and A. {Tongen}},
booktitle={The American Mathematical Monthly},
title={Solitaire Mancala Games and the Chinese Remainder Theorem},
year={2013},
volume={},
number={},
pages={706-724},
}

@INPROCEEDINGS{Tesauro95acm,
author={G. {Tesauro }},
booktitle={Communications of the ACM},
title={Temporal Difference Learning and TD-Gammon },
year={1995},
volume={38}
}

@INPROCEEDINGS{Ferrucci12tiw,
author={D. A. Ferrucci},
booktitle={IBM Journal of Research and Development},
title={Introduction to “This is Watson"},
year={2012},
volume={56},
pages={1:1-1:15}
}

@INPROCEEDINGS{Browne18mtag,
author={C. Browne},
booktitle={2018 IEEE Conf. Comput. Intell. Games},
title={Modern techniques for ancient games},
year={2018},
volume={},
number={},
pages={490–497},
}

@INPROCEEDINGS{Piette19ludii,
author={E. Piette and D. J. N. J. Soemers and M. Stephenson and C. F. Sironi and M. H. M. ´ Winands and C. Browne},
booktitle={CoRR},
title={Ludii - the ludemic general game system},
year={2019},
volume={abs/1905.05013},
number={},
pages={},
}

@InProceedings(IJCAI07bikram,
        author="Bikramjit Banerjee and Peter Stone",
	title="General Game Learning using Knowledge Transfer",
	BookTitle="The 20th International Joint Conference on Artificial Intelligence",
	year={2007},
	pages="672--677",
	wwwnote={<a href="http://www.ijcai-07.org/">IJCAI-07</a>},
)

@INPROCEEDINGS{Piette19ludiiGGSintro,
author={E. Piette and D. J. N. J. Soemers and M. Stephenson and C. F. Sironi and M. H. M. Winands and C. Browne},
booktitle={Proceedings of Advances in Computer Games (ACG 2019),Macao, Springer},
title={A Practical Introduction to the Ludii General Game System},
year={2019},
volume={},
number={},
pages={},
}

@INPROCEEDINGS{Ferns04MDP,
author={N. Ferns and P. Panangaden and D. Precupand},
booktitle={The Nineteenth National Conference on Artificial Intelligence, Association for the Advancement of Artificial Intelligence},
title={Metrics for Finite Markov Decision Processes},
year={2004},
volume={},
number={},
pages={104-124},
}

@inproceedings{yang11,
author = {Cai, Yang and Daskalakis, Constantinos},
title = {On Minmax Theorems for Multiplayer Games},
publisher = {Society for Industrial and Applied Mathematics},
address = {USA},
booktitle = {Proceedings of the Twenty-Second Annual ACM-SIAM Symposium on Discrete Algorithms},
pages = {217–234},
numpages = {18},
location = {San Francisco, California},
year = {2011},
series = {SODA '11}
}

@INPROCEEDINGS{Lorentz_ml,
  title={Machine Learning in the Game of Breakthrough},
  author={R. Lorentz and IV TeofiloErinZosa},
  booktitle={ACG},
  year={2017}
}

@INPROCEEDINGS{Wang18mcqlfggp,
author={H. Wang and M. Emmerich and A. Plaat},
booktitle={	arXiv:1802.05944 [cs.AI]},
title={Monte Carlo Q-learning for General Game Playing},
year={2018},
volume={},
number={},
pages={},
}

@INPROCEEDINGS{evendar03a,
author={E. E.Dar, Y. Mansour},
booktitle={Journal of Machine Learning Research},
title={Learning Rates for Q-learning},
year={2003},
volume={5},
number={},
pages={},
}

@INPROCEEDINGS{Mahadevan94mlp,
author={S. Mahadevan},
booktitle={Machine Learning Proceedings, Proceedings of the Eleventh International Conference, Rutgers University, New Brunswick, NJ},
title={To Discount or not to Discount in Reinforcement Learning: A Case Study Comparing R Learning and Q-learning},
year={1994},
volume={},
number={},
pages={164-172},
}

@article{KIUMARSI20141167,
title = {Reinforcement Q-learning for optimal tracking control of linear discrete-time systems with unknown dynamics},
journal = {Automatica},
volume = {50},
number = {4},
pages = {1167-1175},
year = {2014},
issn = {0005-1098},
doi = {https://doi.org/10.1016/j.automatica.2014.02.015},
url = {https://www.sciencedirect.com/science/article/pii/S0005109814000533},
author = {Bahare Kiumarsi and Frank L. Lewis and Hamidreza Modares and Ali Karimpour and Mohammad-Bagher Naghibi-Sistani},
}

@incollection{MAHADEVAN1994164,
title = {To Discount or not to Discount in Reinforcement Learning: A Case Study Comparing R Learning and Q Learning},
editor = {William W. Cohen and Haym Hirsh},
booktitle = {Machine Learning Proceedings 1994},
publisher = {Morgan Kaufmann},
address = {San Francisco (CA)},
pages = {164-172},
year = {1994},
isbn = {978-1-55860-335-6},
doi = {https://doi.org/10.1016/B978-1-55860-335-6.50028-3},
url = {https://www.sciencedirect.com/science/article/pii/B9781558603356500283},
author = {Sridhar Mahadevan},
}

@ARTICLE{7080922,  author={Gévay, Gábor E. and Danner, Gábor},  journal={IEEE Transactions on Computational Intelligence and AI in Games},   title={Calculating Ultrastrong and Extended Solutions for Nine Men’s Morris, Morabaraba, and Lasker Morris},   
year={2016},  
volume={8},  
number={3},
pages={256-267},  
doi={10.1109/TCIAIG.2015.2420191}
}

@INPROCEEDINGS{Hagen00q,
    author = {Stephan Ten Hagen and Ben Kröse},
    title = {Q-Learning for systems with continuous state and action spaces.},
    booktitle = {In BENELEARN 2000, 10th Belgian-Dutch Conference on Machine Learning},
    year = {2000}
}

@inbook{doi:10.1137/1.9781611973082.20,
author = {Yang Cai and Constantinos Daskalakis},
title = {On Minmax Theorems for Multiplayer Games},
booktitle = {Proceedings of the 2011 Annual ACM-SIAM Symposium on Discrete Algorithms (SODA)},
year = 2011,
chapter = {},
pages = {217-234},
doi = {10.1137/1.9781611973082.20},
URL = {https://epubs.siam.org/doi/abs/10.1137/1.9781611973082.20},
eprint = {https://epubs.siam.org/doi/pdf/10.1137/1.9781611973082.20}
}

@INPROCEEDINGS{6633628,  
author={Divilly, Colin and O'Riordan, Colm and Hill, Seamus},  
booktitle={2013 IEEE Conference on Computational Inteligence in Games (CIG)},   title={Exploration and analysis of the evolution of strategies for Mancala variants},  
year={2013},  
volume={},  
number={},  
pages={1-7},  
doi={10.1109/CIG.2013.6633628}
}

@article{VANDENHERIK2002277,
title = {Games solved: Now and in the future},
journal = {Artificial Intelligence},
volume = {134},
number = {1},
pages = {277-311},
year = {2002},
issn = {0004-3702},
doi = {https://doi.org/10.1016/S0004-3702(01)00152-7},
url = {https://www.sciencedirect.com/science/article/pii/S0004370201001527},
author = {H.Jaap {van den Herik} and Jos W.H.M. Uiterwijk and Jack {van Rijswijck}},
}

@INPROCEEDINGS{GassersSNMM96,
author={R. Gasser},
booktitle={Computational Intelligence},
title={Solving Nine Men's Morris},
year={1996},
volume={12},
pages={24-41},
}

@INPROCEEDINGS{Volz20Benchmark,
  author={V. {Volz} and B. {Naujoks}},
  booktitle={arXiv:2007.02742v1 [cs.AI]}, 
  title={Towards Game-Playing AI Benchmarks via Performance Reporting Standards}, 
  year={2020}}

@article{schaeffer2007checkers,
  title={Checkers is solved},
  author={Schaeffer, Jonathan and Burch, Neil and Bjornsson, Yngvi and Kishimoto, Akihiro and Muller, Martin and Lake, Robert and Lu, Paul and Sutphen, Steve},
  journal={science},
  volume={317},
  number={5844},
  pages={1518--1522},
  year={2007},
  publisher={American Association for the Advancement of Science}
}

@article{van2002games,
  title={Games solved: Now and in the future},
  author={Van Den Herik, H Jaap and Uiterwijk, Jos WHM and Van Rijswijck, Jack},
  journal={Artificial Intelligence},
  volume={134},
  number={1-2},
  pages={277--311},
  year={2002},
  publisher={Elsevier}
}

@article{allen1989note,
  title={A note on the computer solution of Connect-Four},
  author={Allen, J},
  journal={The First Computer Olympiad},
  pages={134--135},
  year={1989},
  publisher={Ellis Horwood}
}

\end{document}